\documentclass[11pt]{article}
\usepackage{eamt18}
\usepackage{times}
\usepackage{url}
\usepackage{latexsym}
\usepackage[small,bf]{caption} %
\usepackage{amsmath,amssymb}
\usepackage{multirow}
\usepackage[english]{babel}
\usepackage{xspace}
\usepackage{graphicx}
\usepackage{array}
\newcolumntype{H}{>{\setbox0=\hbox\bgroup}c<{\egroup}@{}}

\usepackage{tikz}
\usetikzlibrary{shapes,arrows}
\usetikzlibrary{calc, positioning}

\DeclareMathOperator*{\argmax}{arg\,max}

\newcommand\tuple[1]{\langle #1 \rangle}

\tikzstyle{io} = [trapezium,trapezium left angle=70,trapezium right angle=-70,minimum height=3em, draw, fill=blue!20, 
text width=4.5em, text badly centered,inner sep=0pt]
\tikzstyle{decision} = [diamond, draw, fill=blue!20, 
    text width=4.5em, text badly centered, inner sep=0pt]
\tikzstyle{block} = [rectangle, draw, fill=blue!20, 
    text width=5em, text centered, rounded corners, minimum height=3em]
\tikzstyle{line} = [draw, -latex']
\tikzstyle{cloud} = [draw, ellipse,fill=red!20, node distance=3cm,
    minimum height=2em]
\newcommand\esen{Spanish$\rightarrow$English\xspace}
\newcommand\enes{English$\rightarrow$Spanish\xspace}

\title{Bootstrapping Multilingual Intent Models\\ 
via Machine Translation for Dialog Automation}

\author{Nicholas Ruiz, Srinivas Bangalore, John Chen \\
         Interactions, LLC \\ 
         Murray Hill, NJ, USA \\
         \{\texttt{nruiz,sbangalore,jchen}\}@interactions.com}

\date{}

\begin{document}
\maketitle
\begin{abstract}
With the resurgence of chat-based dialog systems in consumer and enterprise applications, there has been much success in developing data-driven and rule-based natural language models to understand human intent.
Since these models require large amounts of data and in-domain knowledge, expanding an equivalent service into new markets is disrupted by language barriers that inhibit dialog automation.

This paper presents a user study to evaluate the utility of out-of-the-box machine translation technology to (1) rapidly bootstrap multilingual spoken dialog systems and (2) enable existing human analysts to understand foreign language utterances.
We additionally evaluate the utility of machine translation in human assisted environments, where a portion of the traffic is processed by analysts.
In \enes experiments, we observe a high potential for dialog automation, as well as the potential for human analysts to process foreign language utterances with high accuracy.
\end{abstract}

\section{Introduction}
With the present advances in natural language understanding  and speech recognition technologies, unprecedented opportunities have been created for realizing natural and sophisticated human-machine conversations to accomplish routine tasks. There has been a resurgence of speech/text-based conversation systems spanning multiple platforms, such as interactive voice recordings, chat, and SMS, owing to the availability of communication platforms that make it convenient to configure human-machine conversations.
The potential opportunities of speech/text driven human-machine systems, or \textit{virtual agents}, can only be realized if the user's requests are \textit{understood} by the virtual agent and acted upon appropriately.

For practical applications, such conversational agents and speech/text analytics systems, the meaning of a sentence may be approximated as one or more actionable labels, or \textit{intents}, associated with the input utterance. In such cases, the natural language understanding (NLU) task is  modeled as an \textit{intent classification} problem.
Although ambiguity is present in natural language, 
data-driven NLU systems have been successful in modeling user intents in many application domains.

Many commercial and enterprise applications service customers from different geographic locations and varying language proficiencies, requiring multilingual NLU for human-machine interaction. In order to deploy an intent classification system for a new language
 a new set of labeled training data is conventionally required. 
This data is often unavailable before a solution is deployed, instead requiring a human-driven dialog system depending on intent analysts or live agents. In time, a sufficient amount of production data may be collected to build a data-driven intent model;
however this approach is expensive to operate and ignores valuable knowledge present in other languages that could be used to build an initial model.

In this paper, we evaluate the use of machine translation (MT) as a tool to bridge the knowledge present in one or more intent models for the creation of an intent classifier in a target language. Given MT's capability to translate the content of utterances in a source language to a target language, our goal is to minimize the number of language proficient intent analysts needed to support a production-scale multilingual dialog system in the absence of target language data. 

The remainder of the paper is organized as follows: in Section \ref{sec:nlu-care} we discuss the details of the intent classification system and present the NLU model, including our two MT architectures in a multilingual spoken dialog system. In Section \ref{sec:experiment} we outline our experiment and describe our ASR, MT, and NLU models. 
In Sections \ref{sec:asr-eval}-\ref{sec:boot-eval}, we evaluate the ASR, MT, and NLU model performance.
In Section \ref{sec:agent-eval} we evaluate human agents' ability to label the intents of translated user utterances and summarize our findings in Section \ref{sec:conclusion}.

\section{NLU for Customer Care}
\label{sec:nlu-care}
A single utterance is tagged with three types of labels: \textit{intents}, \textit{entities}, and \textit{conversational handlers}. Intents are domain-specific labels such as \textsc{sales}, \textsc{tech assistance}, and \textsc{billing}.
\textit{Entity} labels represent the names of products or services mentioned by the user. These include specific models of smart phones or subscription services.
Conversational handlers are labels which are similar to speech acts to guide the conversation. For example, \textsc{live agent}, \textsc{confused}, or \textsc{foreign language}.
In our experiments, intents and entities are labeled as ``session variables'' (SV), while conversation handles are partitioned into ``task names'' (TN) and ``event names'' (EN).
Examples of each in our experiments are shown in Table~\ref{tbl:intent-labels}.

A joint SVM classifier is trained by concatenating the TN, EN, and SV labels into a unique label. The model comprises a set of binary SVM classifiers, with each classifier predicting if the input is assigned or not assigned to a particular label type.
For a given input utterance $x$, the joint label is computed as:
\begin{equation}
y^{*} = \argmax_{y \in \tuple{tn,sv,en}} F_{y}(x,y).
\end{equation}
The feature set \textit{F} may comprise \textit{application context}, \textit{conversation context}, and the \textit{utterance}. We use $n$-gram word-level features in this experiment.

\begin{table}[bt]
\centering
\footnotesize
\resizebox{1.0\linewidth}{!}{%
\begin{tabular}{|l|l|l|}
\hline
       Task Names (TN) 	&	       Session Variables (SV)  	&	       Event Names     (EN) \\
                \hline                                  				
       COMPLAINT       	&	       ACCOUNT {action}	&	       ANGRY   \\
       ENGLISH 	&	       ACTIVATE {product}      	&	       DON'T KNOW      \\
       FOREIGN        	&	       ADD {service}   	&	       GARBLED \\
       NONE    	&	       APPOINTMENT {type}      	&	       LIVE AGENT      \\
       QUESTION        	&	       BILLING AUTO PAY        	&	       NOISE   \\
       ...     	&	       BILLING DETAILS 	&	       NO MATCH        \\
               	&	       PAY BILL {service}      	&	       NONE    \\
               	&	       CHANGE ADDRESS  	&	       THANK YOU       \\
               	&	       DISCONNECT {service}    	&	   ...     \\
               	&	       ... 	&	\\
\hline
10	&	707	&	18 \\			
\hline
\end{tabular}%
}
\caption{Examples and counts of Spanish intent labels by category and language for the ``How may I help you?'' dialog state. Concatenating the labels yields 833 distinct intent annotations.}
\label{tbl:intent-labels}
\end{table}

\subsection{Confidence Measures}
We boost the accuracy of our intent models by using human analysts' predictions on unconfident decisions made by the classifier. We obtain prediction probabilities from the classifier by computing the sigmoid on the scores output for each label by the SVM classifier, computed as:
\begin{equation}
P(y^{*}) = \frac{1}{1 + \exp(F_{y}(x, y^{*}))}.
\label{eq:sigmoid}
\end{equation}
The confidence measure is obtained by computing the ratio of the probabilities of the first and second best labels assigned by the classifier:
\begin{equation}
cf(y^{*}) = \frac{P(y^{*})}{P(y^{*-1})}.
\label{eq:confidence-nl}
\end{equation}
The rejection threshold is empirically determined to maximize the accuracy of the human-assisted solution while minimizing human labeling costs.

\begin{figure}[tb]
\centering
\resizebox*{0.58\width}{0.47\height}{%
\begin{tikzpicture}[node distance = 2cm, auto]
    \node [io] (init) {Spanish audio};
    \node [block, below of=init] (asr) {Spanish ASR};
    \node [decision, below of=asr, node distance=2cm] (asr_ok) {ASR OK?};
    \node [block, below of=asr_ok] (mt) {Spanish-English MT};
    \node [decision, below of=mt] (mt_ok) {MT OK?};
    \node [block, right of=mt_ok, node distance=4.5cm] (spanish_ia) {Spanish IA};
    \node [block, below of=mt_ok] (nl) {English NL};
    \node [decision, below of=nl] (nl_ok) {NL OK?};
    \node [block, right of=nl_ok, node distance=3cm] (english_ia) {English IA};
    \node [io, below of=nl_ok, node distance=2.5cm] (intent) {Spanish Intent};
    \path [line] (init) -- (asr);
    \path [line] (asr) -- (asr_ok);
    \path [line] (asr_ok) -- node {yes} (mt);
    \path [line] (mt) -- (mt_ok);
    \path [line] (mt_ok) -- node {yes} (nl);
    \path [line] (nl) -- (nl_ok);
    \path [line] (asr_ok) -| node {no} (spanish_ia);
    \path [line] (mt_ok) -- node {no} (spanish_ia);
    \path [line] (nl_ok) -- node {no} (english_ia);
    
    \path [line] (nl_ok) -- node {yes} (intent);
    \path [line] (spanish_ia) |- ++(0,-5.1cm) -| (intent.north);
    \path [line] (english_ia) |- ++(0,-1.1cm) -| (intent.north);
\end{tikzpicture}%
}
\caption{Online \esen bootstrap architecture. Spanish audio is translated into English and processed by an English intent classifier. If ASR or MT scores are not confident, a Spanish intent analyst (IA) labels the segment. If the English intent model is not confident, an English IA labels it.}
\label{fig:online-mt}
\end{figure}
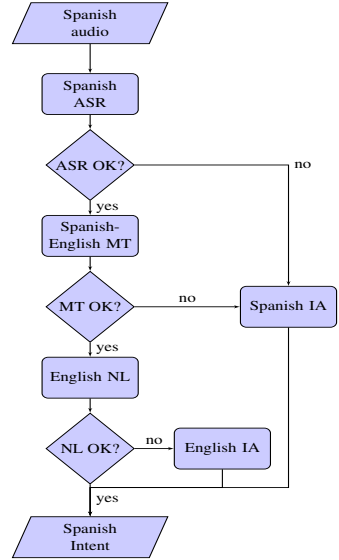

\subsection{Multilingual Bridging via Machine Translation}
\label{sec:mt-nlu}
In the context of dialog systems, MT can be used in one of two ways: (1) translating real-time target language data into the source language and predicting the intent with a source language intent classification model (cf. Fig.~\ref{fig:online-mt}); or (2) translating source language data offline into the target language and training a target language intent classification model. 
In the first scenario, utterances with high translation quality may be processed by an analyst that only speaks the source language, if the NLU confidence score is too low. 

\section{Experimental setup}
\label{sec:experiment}
We evaluate the efficacy of bootstrapping a Spanish intent classifier using the data and underlying models from an English spoken language dialog system. 
The data set consists of customer voice responses to the message ``How may I help you?'' in the customer's native language at the beginning of a phone call to an Interactive Voice Response (IVR) system.
We assume that no Spanish intent data is available during training time and evaluate the performance of our bootstrapped Spanish models against an intent model trained with a standard training set.
Table \ref{tbl:data-counts} lists the data used in our experiment, 
The training data consists of 5-best ASR hypotheses on audio segments for English and Spanish.
We assume that the intent labels covered by the target Spanish model are the same as the English model. Although most of the intent labels overlap one another there is a subset of $\tuple{tn,sv,en}$ intent triples that do not overlap. We discard the non-training examples with non-overlapping triples from each data set, reducing the number of unique labels in both data sets to 623 (cf. Table \ref{tbl:intent-labels}). 
As a result, 2.94\% of the English training examples and 1.95\% of the Spanish training examples are discarded, respectively.
\begin{table}[bt]
\centering
\footnotesize
\resizebox{1.0\linewidth}{!}{%
\begin{tabular}{lrrr}
\hline\hline
Data set	&	\# utts&	\# words	 & \# unique labels \\
\hline
English training	&	6.5M	&	40.6M		& 	623 \\
Spanish training	&	0.8M	&	4.6M	& 	623 \\
Spanish test (ASR)	&	1007	&	4696		& 	178 \\
Spanish test (human)	&	1007	&	5153		&	178 \\
\hline
\end{tabular}%
}
\caption{Utterance, word, and distinct label counts for the training and test data used in our experiments. 
}
\label{tbl:data-counts}
\end{table}
Model performance is evaluated on a test set of 1007 Spanish audio segments randomly extracted from production logs and labeled by Spanish-speaking intent analysts (IAs) and verified by a supervisor.

\subsection{Automatic Speech Recognition}
Our Spanish ASR system consists of an n-gram language model and hybrid DNN acoustic model trained with the cross-entropy criterion followed by the state-level Minimum Bayes Risk (sMBR) objective. We use sequence-training with smoothing and speed-perturb the training data. The acoustic model has general-phone and head-body-tail based digit-specific triphones. The training data consists of 500K training utterances (around 1000 hours of audio) without speech perturbations and about 40K unique vocabulary words.

\subsection{Machine Translation}
We use a conventional neural machine translation (NMT) sequence-to-sequence encoder-decoder with attention architecture \cite{Bahdanau:15,Luong:15:IWSLT,Sennrich:16a} commonly used by MT practitioners.
The NMT models were trained with parallel English-Spanish data from Europarl v7, CommonCrawl, and WMT News Commentary v8 from the WMT 2013 evaluation campaign \cite{bojar-EtAl:2013:WMT}, as well as the TED talks from IWSLT 2014 \cite{Cettolo:14:IWSLT}.
The training data has a shared vocabulary size of 89,500 words after byte-pair encoding \cite{Sennrich:16a}. The model is trained for 20 epochs with two bidirectional LSTM encoding and decoding layers with 512 units.
In this experiment we assume to have no in-domain parallel data.%

For the offline \enes model, we translate the English intent model's training data (6.5 million utterances) into Spanish using our baseline NMT system.
The number of words in the translated data set remains roughly the same.
The translated outputs are used to train a bootstrapped Spanish intent classifier, using the same training parameters as the native English model. The ASR outputs from the test set are processed by the bootstrapped Spanish intent classifier.
For the real-time \esen model, insert punctuation and apply truecasing to the ASR outputs from the test set and translate the outputs with our \esen baseline NMT system. We strip the punctuation and lowercase the machine translated output and pass it through the native English intent classifier.

\subsection{Intent classification}
Our intent classifiers are trained using an implementation of SVMs in \textsc{scikit-learn}\footnote{http://scikit-learn.org}, using the approach described in Section \ref{sec:nlu-care}. 
We evaluate the performance of an intent classification model by plotting an \textit{error-rejection curve}, which measures the error rate of the intent classifier as the number of utterances that are processed by the intent analysts increases. For example, a 10\% rejection rate corresponds means that only 90\% of the test set is evaluated by the model.

\begin{figure}[bt]
\centering
\vspace{-0.25cm}
\includegraphics[width=1.1\linewidth]{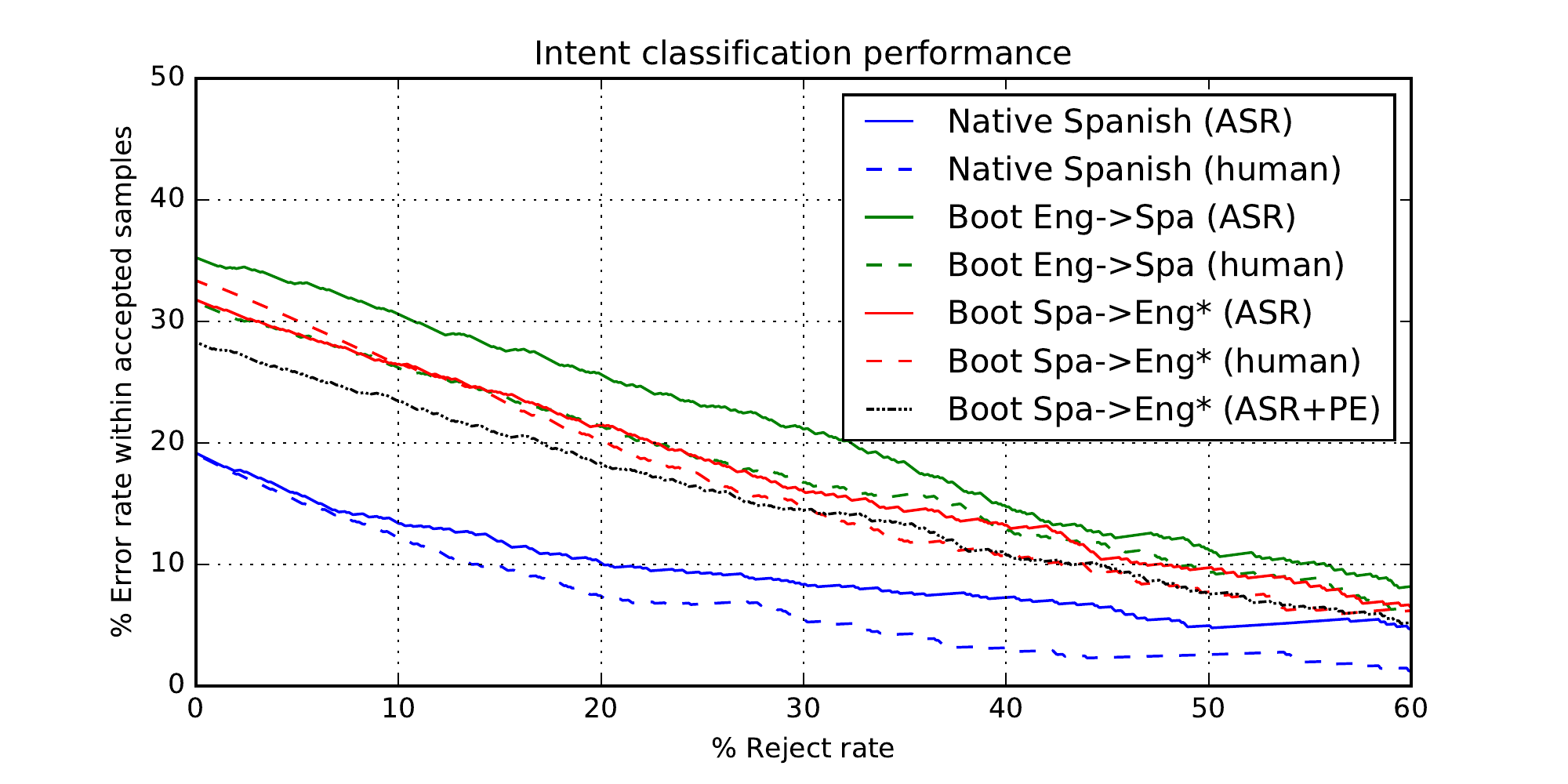}
\caption{Error-rejection curves for bootstrapped intent classification performance on Spanish ASR outputs, versus a native Spanish model. Results are reported for intent classifier predictions on ASR outputs.
Spa$\rightarrow$Eng**: performance on post-edited MT outputs.
}
\label{fig:err-intent}
\end{figure}

We compare the results of each bootstrapped NLU approach with a native Spanish intent model trained on the held-out Spanish training data.
We additionally repeat the experiment with the human transcripts to measure the difference in intent classification error that may be explained by ASR.
Error-rejection curves for each intent model are shown in Fig. \ref{fig:err-intent} and the scores at 0\%, 10\%, and 20\% rejection are shown in Table \ref{tbl:err-classify}.

\section{ASR performance}
\label{sec:asr-eval}
We use \textsc{sclite} from the NIST Speech Recognition Scoring Toolkit\footnote{https://www.nist.gov/itl/iad/mig/tools} to compute the word error rate (WER) and utterance error rate.
After further clean-up and adjudication, we observe a 33.2\% WER, with 60.0\% of the utterances containing errors (32\% substitutions, 22\% deletions, 28\% insertions).
A majority of the substitution errors were confusions between singular and plural (e.g. \textit{problemas}$\rightarrow$\textit{problema}) and articles (e.g. \textit{del}$\rightarrow$\textit{de}). Other errors included phonetic confusions (e.g. \textit{cuenta}$\rightarrow$\textit{fuenta}), named entity misrecognitions (e.g. \textit{HBO}$\rightarrow$\textit{yo}), and a high frequency of dropped articles (e.g. \textit{de}, \textit{la}, \textit{a}) caused by speaker under-articulation.
Of these types of errors, the most detrimental are substitutions of named entities, verbs, and nouns that are not lemmatization errors.
Another issue driving up the WER score caused by the IVR system truncating audio longer than four seconds to reduce latency. %

\begin{table}[tb]
\centering
\footnotesize
\begin{tabular}{l|rrr}
\hline
Translation	&	       BLEU $\uparrow$ 	&	       TER $\downarrow$        	&	       Length  \\
\hline
       ASR outputs    	&	42.5	&	51.2	&	       94.1    \\
       ASR outputs (+PE)	&	48.0	&	44.6	&	96.3 \\
       Human transcripts   	&	51.8	&	41.3	&	       111.4   \\
\hline
\end{tabular}
\caption{Machine translation quality measured in BLEU, TER, and utterance length, evaluated against post-edited translations of human transcripts. 
}
\label{tbl:mt-eval}
\end{table}

\section{Machine Translation quality}
\label{sec:mt-eval}
In order to assess the translation quality, we post-edit the translations of the human-transcribed utterances
and report the case-insensitive BLEU and TER scores in Table \ref{tbl:mt-eval}.
ASR errors increase the required translation edits by 10\%, from 41.3\% edits to 51.2\%. 
The primary sources of errors are incomplete sentences, lack of punctuation;
lexical mistranslations of key words: 
(e.g. \textit{equipo}~(\textit{equipment})$\rightarrow$\textit{team}; 
\textit{direcci\'{o}n} (\textit{address})$\rightarrow$\textit{direction};
\textit{reclamo}~(\textit{complaint})$\rightarrow$ \textit{claim});
ambiguous translations 
(e.g. \textit{factura}$\rightarrow$\textit{bill/invoice});
and duplicated words during translation
(e.g. \textit{payment arrangements}$\rightarrow$\textit{payment payment}).
Many of these issues are due to lack of in-domain MT training data and low tolerance of ASR errors.
Highly repetitive errors indicate that an automatic post-editing system could substantially improve the translation system's quality.
Table \ref{tbl:mt-eval} also shows that post-edited translations of ASR outputs are substantially worse than those of the human transcripts (41.3\% TER difference), showing that ASR errors are exacerbated through translation.

\section{Native NLU performance}
\label{sec:native-eval}
Our reference native Spanish intent classification model is trained on ASR outputs since an insufficient amount of human-transcribed intent modeling data is available. 
From Table \ref{tbl:err-classify}, we see that at 0\% rejection, the native model yields a 19.2\% classification error, while the human intent analysts (IAs) yielded a 11.0\% error while listening directly to the audio. At the same time, if the IAs are presented with the ASR outputs, they produce an error rate of 24.4\%. This demonstrates the intent model's ability to tolerate a certain degree of ASR errors by being trained on ASR errors.

Of the 1007 test utterances, 152 utterances have both ASR errors and were classified incorrectly, although they were labeled properly by IAs, comprising 15.1\% of the 19.2\% intent classification errors. 
Of the utterances with ASR errors that cause an intent classification error, six were cases where ASR failed to produce a hypothesis, 26 were cases of audio truncation, and 39 utterances containing only ASR substitution errors. The latter cases often caused underspecification during intent classification 
(e.g. \textsc{lower my bill}$\rightarrow$\textsc{billing problem}; \textsc{make a payment}$\rightarrow$\textsc{billing}).
In isolation, insertion and deletion ASR errors did not have a significant impact on NLU.

Evaluating NLU performance on human transcripts, we observe improvements that rival human labeling performance at 10\% rejection and above. At 20\% rejection, the model significantly outperforms human labeling.

\begin{table}[bt]
\centering
\footnotesize
\begin{tabular}{l|rrr}
\hline\hline
						&   \multicolumn{3}{c}{Reject}\\
Configuration			&	0\% 	&	10\% &	20\% \\
\hline
Native	Spanish	(ASR)	&	19.2	&	13.5	&	10.3	\\
Native	Spanish	(human)	&	19.1	&	12.3	&	7.4	\\
Boot 	Eng$\rightarrow$Spa	(ASR)	&	35.3	&	30.7	&	25.8	\\
Boot	 Eng$\rightarrow$Spa	(human)	&	31.6	&	26.3	&	21.4	\\
Boot 	Spa$\rightarrow$Eng*	(ASR)	&	31.8	&	26.5	&	21.4	\\
Boot 	Spa$\rightarrow$Eng*	(human)	&	33.4	&	26.7	&	20.2	\\
Boot	 Spa$\rightarrow$Eng*	(ASR+PE)	&	28.3	&	23.6	&	18.3	\\
	\hline								
Spanish	IA	(audio)	&	11.0	&	10.6	&	9.5	\\
Spanish IA (human) & 14.9 & 13.1 & 10.9 \\
Spanish	IA	(ASR)	&	24.4	&	21.6	&	17.6	\\
English	IA	(ASR+MT)	&	33.9	&	30.6	&	26.1	\\
English	IA	(ASR+MT+PE)	&	25.9	&	22.8	&	18.3	\\
\hline
\end{tabular}
\caption{Intent classification performance by machine learning models and human intent analysts (IAs) at 0\%, 10\%, and 20\% rejection.}
\label{tbl:err-classify}
\end{table}

\section{Bootstrapped NLU performance}
\label{sec:boot-eval}
While the native Spanish model has a 13.5\% error rate at 10\% rejection when processing ASR hypotheses, the bootstrapped models have double the error rate due to the use of out-of-domain machine translation. On ASR hypotheses, the \enes model yields an error rate of 30.7\%, while the \esen model yields 26.5\% at 10\% rejection.

To better understand how machine translation further corrupts the bootstrapped \esen performance, we compare the errors it makes to the Native Spanish model. In Table \ref{tbl:esen-compare}, we group the NLU errors by whether they are present in the Native Spanish model, the bootstrapped model, or both.
14.2\% of the test set are examples where the Native Spanish model makes a correct prediction, but the \esen model yields errors. By comparing to post-edited ASR+MT data, only 3\% of those errors are directly attributed to ASR errors.
The 11\% of \esen model-specific errors are mostly attributed to intent underspecification. 
For example, 60\% of the \textsc{account} errors are \textsc{null} misclassifications.
For billing issues, two-thirds of the errors are semantically similar misclassifications, such as \textsc{payment}, \textsc{lower my bill}, and \textsc{bill details}.

\begin{table}[bt]
\centering
\footnotesize
\begin{tabular}{ll|rrH}
\hline\hline
Native	&	Spa$\rightarrow$Eng	&	ASR	&	ASR+PE	&	Human	\\
\hline
+	&	+	&	66.6\%	&	69.6\%	&	62.3\%	\\
+	&	-	&	14.2\%	&	11.2\%	&	18.6\%	\\
-	&	+	&	1.5\%	&	2.0\%	&	4.4\%	\\
-	&	-	&	17.7\%	&	17.2\%	&	14.8\%	\\
\hline
\end{tabular}
\caption{Comparison of Native Spanish intent model to bootstrapped \esen models on ASR outputs.
+/- indicate that whether the corresponding model's prediction was correct.}
\label{tbl:esen-compare}
\end{table}

\section{Analyst performance on translated text}
\label{sec:agent-eval}
\begin{figure}[bt]
\centering
\vspace{-0.25cm}
\includegraphics[width=1.1\linewidth]{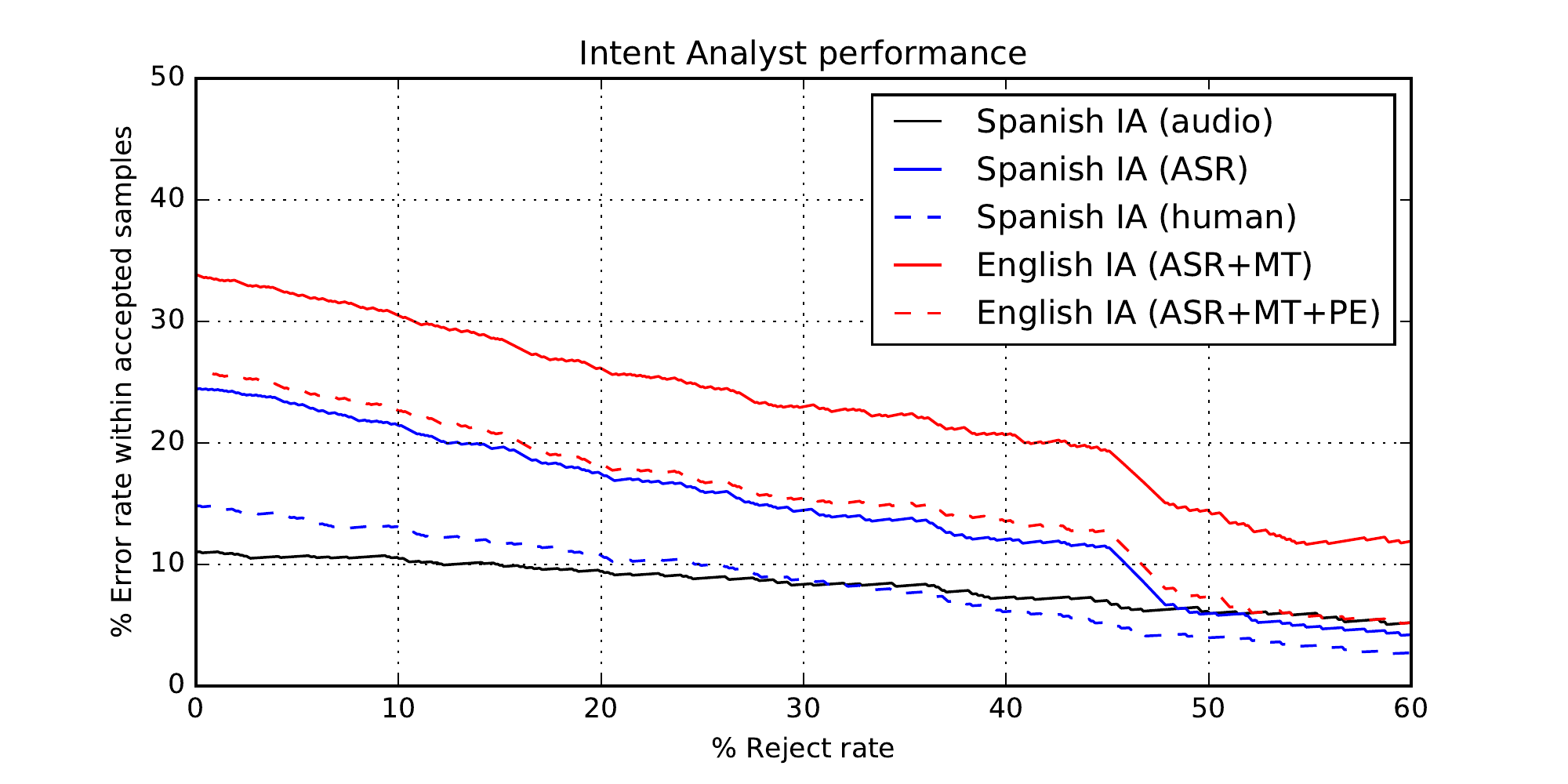}
\caption{Error-rejection curves for the intent analyst (IA) labeling accuracy on Spanish audio, Spanish ASR and human transcripts, and their machine translations into English (MT). 
Rejection is computed from the English model's confidence scores on MT outputs.
``+PE'': performance on post-edited MT outputs.
}
\label{fig:err-analyst}
\end{figure}

Finally, we measure IA labeling performance on translated utterances. 
In our conventional scenario, intent analysts listen to audio segments in their native language and provide an intent label.
Instead, we replace the original audio with machine translated or translation post-edits of ASR hypotheses.
Fig.~\ref{fig:err-analyst} provides error-rejection curves for IAs, with error rates at 0\%, 10\% and 20\% rejection reported in Table \ref{tbl:err-classify}.

We first assess the labeling loss when humans annotate ASR outputs in the absence of audio.
Although their error rate increases from 11.0\% to 24.4\% when annotating ASR transcripts, their performance on human transcripts is within 5\% of listening directly to the audio at 0\% rejection. As the rejection rate increases, the difference becomes negligible.
As we introduce \esen machine translation, we observe that the labeling error increases from 24.4\% to 33.9\% on ASR, which is incidentally worse than the \esen intent classification model's accuracy (31.8\%)! 
However, the IA labeling error rate drops to 25.9\% on post-edited MT outputs
-- only a 1.5\% increase in NLU errors caused by translation. These results suggest that with proper ASR and MT adaptation through in-domain data, we could obtain similar English-speaking IA performance on machine translation outputs as the Spanish-speaking IAs on their native language utterances.

\section{Related Work}
\label{sec:related}
The use of MT to translate texts in other languages into English for sentiment analysis was proposed in Denecke~\shortcite{denecke:08:mt-sentiment}.
Bautin et al.~\shortcite{conf/icwsm/BautinVS08} show that sometimes MT performs inadequate translations on essential parts of a text, affecting sentiment analysis performance. Our results confirm this phenomena due to a lack of in-domain MT training data.
Schwenk and Douze~\shortcite{Schwenk:W17-2619} explore learning multilingual sentence embeddings with neural MT, which can aid in multilingual search.
Prior to that, multilingual approaches leveraged lexical resources such as MultiWordNet \cite{MultiWordNet_2002} to bridge concepts from one language to another.

\section{Conclusions}
\label{sec:conclusion}
We have executed an experiment to measure machine translation's ability to rapidly bootstrap intent classification models for new languages. In our \enes experiments, we observe that although the initial results appear to be substantially worse than a Native Spanish intent classification model, we show that MT can provide a degree of automation that supports human-assisted multilingual dialog systems that can be deployed to production on day one, reducing the need for human agent support over a fully manual solution.
There is further promise that model improvements can be obtained by improving the ASR and machine translation models to include in-domain data.
Finally, we observe it is better to use the online \esen bootstrap in our production system rather than an offline \enes intent model.

\bibliography{paper}
\bibliographystyle{eamt18}

\end{document}